\pdfoutput=1

\documentclass[11pt]{article}

\usepackage{ACL2023}

\usepackage{times}
\usepackage{latexsym}

\usepackage[T1]{fontenc}

\usepackage[utf8]{inputenc}

\usepackage{microtype}

\usepackage{inconsolata}

\usepackage{graphicx}  
\usepackage{amsmath,amsfonts,amssymb,amsthm}
\usepackage{multicol,multirow}
\usepackage{algorithm}
\usepackage{algpseudocode}
\usepackage{paralist}
\usepackage{bbm}
\usepackage{booktabs}

\theoremstyle{definition}
\newtheorem{example}{Example}[section]

\newcommand{\Vertex}{\mathcal{V}}
\newcommand{\Triple}{\mathcal{T}}
\newcommand{\Relation}{\mathcal{R}}
\newcommand{\KG}{\mathcal{KG}}
\newcommand{\real}{\mathbb{R}}
\newcommand{\wfr}{\mathcal{WFR}}
\newcommand{\bm}{\mathcal{BM}(\real)}
\newcommand{\bmd}{\mathcal{BM}_d}

\title{Wasserstein-Fisher-Rao Embedding: Logical Query Embeddings\\ with Local Comparison and Global Transport}

\author{Zihao Wang\thanks{$\quad$ Equal Contribution} $^{\dagger}$, Weizhi Fei$^{*\clubsuit}$, Hang Yin$^{\clubsuit}$, Yangqiu Song $^{\dagger}$, Ginny Y. Wong$^{\bigstar}$, Simon See$^{\bigstar}$\\
$\dagger$ CSE, HKUST, HKSAR, China\\
$\clubsuit$ Department of Mathematical Sciences, Tsinghua University, Beijing, China\\
$\bigstar$ NVIDIA AI Technology Center (NVATIC), NVIDIA, Santa Clara, USA \\
\texttt{zwanggc@cse.ust.hk}, \texttt{\{fwz22,hyin-20\}@mails.tsinghua.edu.cn} \\
\texttt{yqsong@cse.ust.hk}, \texttt{\{gwong,ssee\}@nvidia.com}
}

\begin{document}
\maketitle
\begin{abstract}
Answering complex queries on knowledge graphs is important but particularly challenging because of the data incompleteness. Query embedding methods address this issue by learning-based models and simulating logical reasoning with set operators. Previous works focus on specific forms of embeddings, but scoring functions between embeddings are underexplored. In contrast to existing scoring  functions motivated by \textit{local} comparison or \textit{global} transport, this work investigates the \textit{local} and \textit{global} trade-off with unbalanced optimal transport theory. Specifically, we embed sets as bounded measures in $\real$ endowed with a scoring function motivated by the Wasserstein-Fisher-Rao metric. Such a design also facilitates closed-form set operators in the embedding space. Moreover, we introduce a convolution-based algorithm for linear time computation and a block-diagonal kernel to enforce the trade-off. Results show that WFRE can outperform existing query embedding methods on standard datasets, evaluation sets with combinatorially complex queries, and hierarchical knowledge graphs. Ablation study shows that finding a better \textit{local} and \textit{global} trade-off is essential for performance improvement.\footnote{Our implementation can be found at~\url{https://github.com/HKUST-KnowComp/WFRE}.}
\end{abstract}

\section{Introduction}

Knowledge graphs (KGs) store real-world factual knowledge as entity nodes and relational edges~\citep{Miller1995WordNetlexical,Bollacker2008Freebasecollaboratively,Vrandecic2014Wikidatafree}. And they facilitate many downstream tasks~\citep{xiong2017explicit,wang2019kgat,lin2020kgnn}. Notably, answering complex logical queries is an essential way to exploit the knowledge stored in knowledge graphs~\citep{Ren2020Query2boxReasoning,Ren2021LEGOLatent}.

\begin{figure*}[t]
\centering
\includegraphics[width=.95\linewidth]{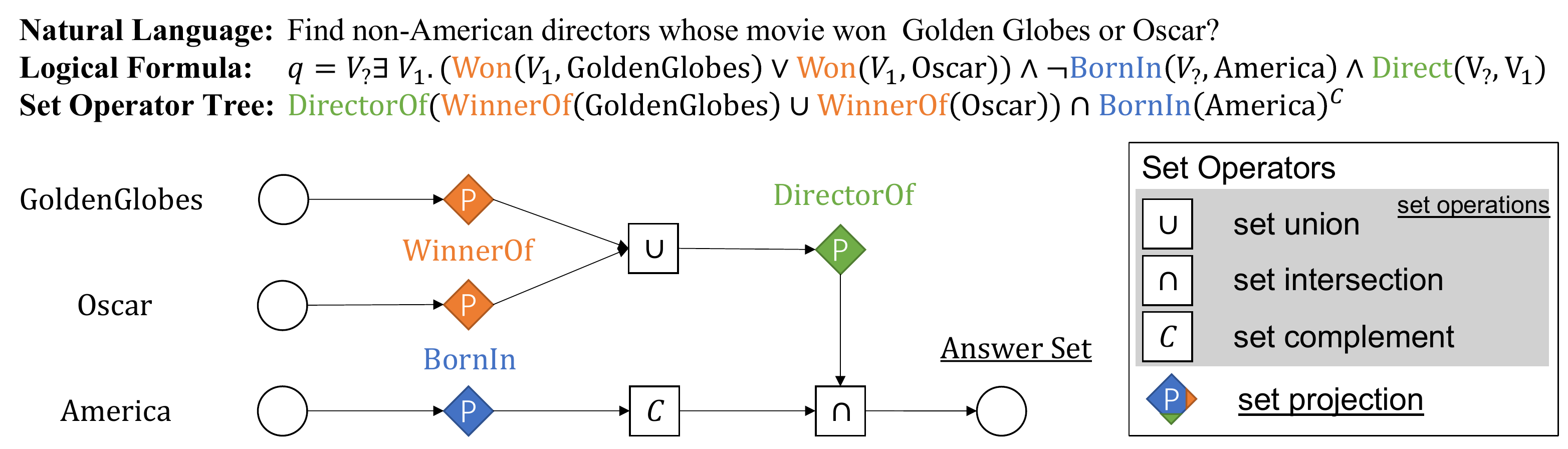}
\caption{Answering logical queries on knowledge graphs. Natural language sentences can be interpreted as logical formulas and then converted to set operator trees~\citep{Wang2021BenchmarkingCombinatorial}.}
\label{fig:example}
\end{figure*}


Formally speaking, complex logic queries can be expressed via first-order logic~\citep{Ren2020Query2boxReasoning,Marker2002Modeltheory}. Specific groups of queries, whose predicates and logical connectives can be converted as set operatiors~\citep{Wang2021BenchmarkingCombinatorial}, are of particular interest due to their clear semantics. Therefore, the logical reasoning process to answering complex queries is transformed to execute set projections and operations in an operator tree~\citep{Ren2020Query2boxReasoning,Wang2021BenchmarkingCombinatorial}. Figure~\ref{fig:example} shows the operator tree for the query "Who is the non-American director that has won Golden Globes or Oscar".

What makes this task difficult is the data incompleteness of knowledge graphs. Modern large-scale KGs are naturally incomplete because they are constructed by crowdsource~\citep{Bollacker2008Freebasecollaboratively,Vrandecic2014Wikidatafree} or automatic information extraction pipelines~\citep{Carlson2010ArchitectureNeverEnding}. This issue is acknowledged as the Open World Assumption~\citep{Libkin2009OpenClosed} (OWA). It leads to the fact that applying query answering algorithms for complete databases will not result in complete answers because of the data incompleteness. Also, it is not able to prune the search space with the observed incomplete knowledge graph, which results in a large computational cost~\citep{Ren2020Query2boxReasoning}. It makes the problem even harder when answering logical queries on large knowledge graphs with billions of edges~\citep{Ren2022SMOREKnowledgea}. We refer readers to recent surveys for more about logical queries on knowledge graphs~\citep{wang2022logical,ren2023neural}.


Query embedding methods~\citep{Hamilton2018EmbeddingLogical,Ren2020Query2boxReasoning} in fixed dimensional spaces are proposed to overcome the above difficulties.
The data incompleteness is addressed by generalizing learnable set embeddings and operators to unseen data~\citep{Ren2020Query2boxReasoning,Ren2020BetaEmbeddings}.
And the computation cost does not grow with the 
Developing efficient forms of set embeddings and operators is one of the recent focuses~\citep{Choudhary2021ProbabilisticEntity,Zhang2021ConECone,Alivanistos2022QueryEmbedding,Bai2022Query2ParticlesKnowledge,Chen2022FuzzyLogic,Yang2022GammaEGamma}.
However, the scoring function between sets, though it also characterizes set embeddings and plays a vital role in training models, is underexplored in the existing literature.
Existing scoring functions are chosen from two categories that emphasize either \textit{local} comparisons~\citep{Ren2020BetaEmbeddings,Amayuelas2022NeuralMethods} or \textit{global} transport  between geometric regions~\citep{Ren2020Query2boxReasoning,Choudhary2021SelfSupervisedHyperboloidc,Zhang2021ConECone}. The following example motivated us to develop scoring functions for embeddings with both \textit{local} and \textit{global} trade-off.

\begin{example}
Consider four "one-hot" vectors with dimension $d=100$:
\begin{align}
    A &= [1,0,0,...,0], \\
    B &= [0,1,0,...,0], \\
    C &= [0,0,1,0,...,0], \\
    D &= [0, …, 0,1].
\end{align}
We observe that:
\begin{compactitem}
    \item Local function (e.g., Euclidean distance) $L$ CANNOT discriminate different similarities between $A$, $B$, $C$, and $D$. Specifically, $L(A, B) = L(A, C) = L(A, D) = L(B, C) = L(B, D) = L(C, D) = 1$.
    \item Global function (e.g. Wasserstein metric) $G$ CAN discriminate. Specifically, $G(A, B) = 1 < G(A, C) = 2 < G(A, D) = 99$. However, G is risky for optimization. For example, if $G(A, D) + G(A, B)$ appears in the objective function of a batch, $G(A, D)$ will dominate $G(A, B)$ because it is 100 times larger, making the optimization ineffective.
    \item Local and global trade-off function (such as the WFR scoring function proposed in this paper) harnesses this risk by constraining the transport within a window size. Our paper finds that the proper window size is 5, which truncated the transport distances between faraway samples like $A$ and $D$. Then, $WFR(A, D) = 5$, and the optimization is stabilized.
\end{compactitem}
\end{example}

In this paper, we develop a more effective scoring function motivated by the Wasserstein-Fisher-Rao (WFR) metric~\citep{chizat2018interpolating}, which introduces the \textit{local} and \textit{global} trade-off.
We propose to embed sets as Bounded Measures in $\real$, where each set embedding can be discretized as a bounded histogram on uniform grids of size $d$.
This set embedding can be interpreted \textit{locally} so that the set intersection, union, and negation can be easily defined by element-wise fuzzy logic $t$-norms~\citep{Hajek1998MetamathematicsFuzzy}.
We propose an efficient convolution-based algorithm to realize the computation of entropic WFR in $O(d)$ time, and a block diagonal kernel to enforce the \textit{local} and \textit{global} trade-off.
We conduct extensive experiments on large number of datasets: (1) standard complex query answering datasets over three KGs~\citep{Ren2020BetaEmbeddings}, (2) large-scale evaluation set emphasizing the combinatorial generalizability of models in terms of compositional complex queries~\citep{Wang2021BenchmarkingCombinatorial}, and (3) complex queries on a hierarchical knowledge graph~\citep{Huang2022LinELogical}. Ablation studies show that the performance of complex query answering can be significantly improved by choosing a better trade-off between \textit{local} comparison and \textit{global} transport.

\section{Related Works}

We discuss other query embedding methods in fixed dimensions and optimal transport in this section. Other methods for complex query answering are discussed in Appendix~\ref{app:other-methods},

\begin{figure*}[t]
\centering
\includegraphics[width=.95\linewidth]{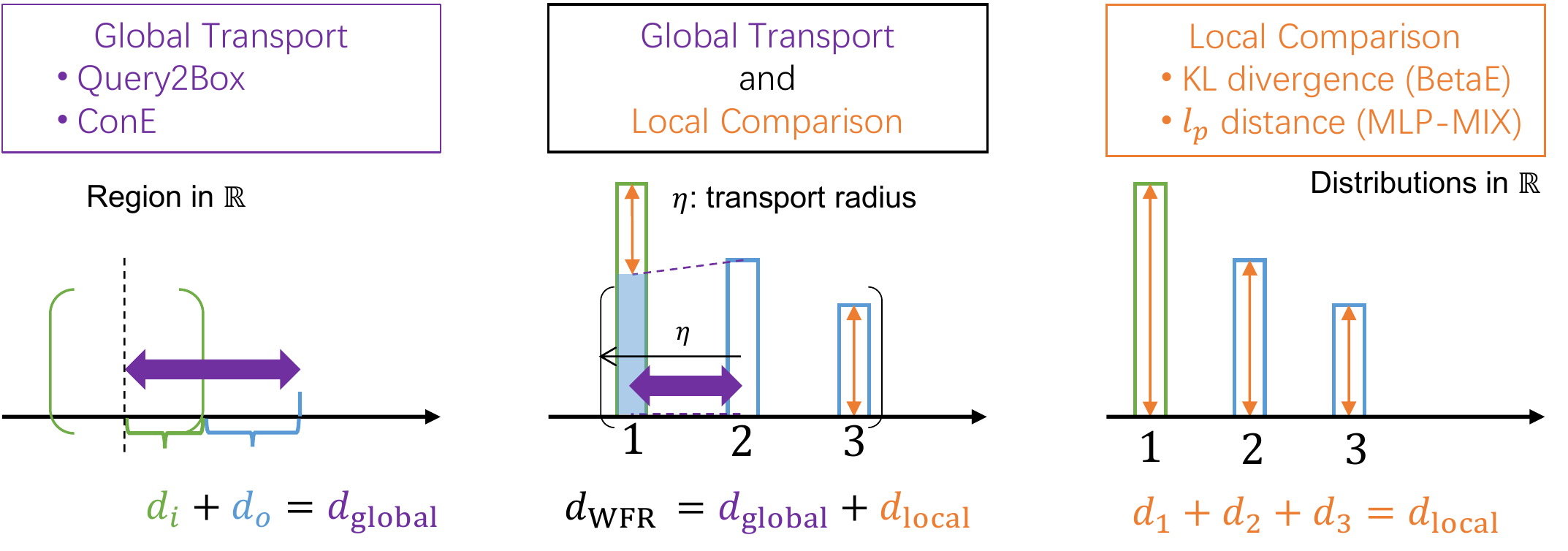}
\caption{Illustration of different scoring functions. \textbf{Left}: global transport, where the difference is measured by how to move mass from one place to another (purple arrows); \textbf{Right}: local comparison, where the difference is measured by in-place comparison (yellow arrows); \textbf{Mid}: local and global trade-off, where we first move mass in the transport radius $\eta$, then compare the unfilled mass.}
\label{fig:local-and-global}
\vskip-.5cm
\end{figure*}

\subsection{Query Embeddings}

As a predominant way to answer logical queries, \textit{query embeddings}~\citep{Hamilton2018EmbeddingLogical} embed answer sets into continuous spaces and models set operations with neural networks. The scope of logical queries that query embedding methods can solve is expanded from conjunctive queries~\citep{Hamilton2018EmbeddingLogical}, to Existential Positive First-Order (EPFO) queries~\citep{Ren2020Query2boxReasoning,Choudhary2021ProbabilisticEntity,Choudhary2021SelfSupervisedHyperboloidc}, and First-Order (FO) queries~\citep{Ren2020BetaEmbeddings,Zhang2021ConECone,Amayuelas2022NeuralMethods,Bai2022Query2ParticlesKnowledge,Yang2022GammaEGamma,wang2023logical,yin2023existential,bai2023sequential}. 

Set embeddings of various forms have been heavily investigated, such as vectors~\citep{Amayuelas2022NeuralMethods,Chen2022FuzzyLogic,Huang2022LinELogical}, geometric regions~\citep{Ren2020Query2boxReasoning,Zhang2021ConECone,Choudhary2021SelfSupervisedHyperboloidc}, and probabilistic distributions~\citep{Ren2020BetaEmbeddings,Choudhary2021ProbabilisticEntity,Yang2022GammaEGamma}. Despite of the various forms of the embeddings, their scoring function captures either \textit{local} comparison, such as Euclidean distance~\citep{Amayuelas2022NeuralMethods}, inner product~\citep{Chen2022FuzzyLogic}, and KL-divergence~\citep{Ren2020BetaEmbeddings,Yang2022GammaEGamma}, or \textit{global} transport, such as heuristic distance between geometric regions~\citep{Ren2020Query2boxReasoning,Choudhary2021SelfSupervisedHyperboloidc,Zhang2021ConECone}, Mahalanobis distance~\citep{Choudhary2021ProbabilisticEntity}, or the similarity between target particle and the closest particle of a point cloud~\citep{Bai2022Query2ParticlesKnowledge}.

In this work, we establish a novel scoring function motivated by unbalanced optimal transport theory~\citep{chizat2018interpolating}. As a variant of the optimal transport, it inherits the advantages and balances the \textit{local} comparison and \textit{global} transport.


\subsection{Optimal Transport for Embeddings} 
Optimal transport (OT)~\citep{peyre2019computational} introduces the power metric between probabilistic distributions and facilitates many applications in language and graph data~\citep{alvarez-melis-jaakkola-2018-gromov,zhao2020semi,xu-etal-2021-vocabulary,li-etal-2021-timeline,tang-etal-2022-otextsum,wang2022unsupervised,li2022gromov,tan2023seal}. It is particularly efficient when embedding graph vertices and words as probabilistic distributions in the Wasserstein space~\citep{muzellec2018generalizing,frogner2018learning}. 

Wasserstein-Fisher-Rao (WFR) metric~\citep{chizat2018interpolating} generalizes the OT between distributions to the measures by balancing local comparison and global transport with a transport radius $\eta$. Existing investigations~\citep{zhao2020relaxed} demonstrated that the WFR metric is a robust and effective measurement for embedding alignment. Previous work measures pretrained embeddings in the WFR
space~\citep{wang2020robust}, while this work is the first to learn embeddings in the WFR space. Moreover, we validate the advantage of WFR space in the context of query embedding.

\section{Preliminaries}
\subsection{Knowledge Graph and Complex Queries}

A knowledge grpah $\KG = \{(h, r, t)\in \Vertex\times\Relation\times\Vertex \}$ is a collections of triples where $h, t\in \Vertex$ are entity nodes and $r\in \Relation$ is the relation.


Complex queries over knowledge graphs can be defined by first-order formulas. Following previous works~\citep{Ren2020BetaEmbeddings}, we consider a query $Q$ with one free variable node $V_?$ and quantified nodes $V_i, 1\leq i\leq n$, an arbitrary logical formula can be converted to prenex and DNF forms as follows~\citep{Marker2002Modeltheory}.
\begin{align}
    Q[V_?] = \square V_1 \cdots \square V_n. c_1 \lor \cdots \lor c_l,
\end{align}
where each quantifier $\square$ is either $\exists$ or $\forall$, each $c_i, 1 \leq i \leq l$ is a conjunctive clause such that $c_i = y_{i1} \land \cdots\land y_{i m_i}$, and each $y_{ij}, 1\leq j\leq m_i$ represents an atomic formula or its negation. That is, $y_{ij} = r(a, b)$ or $\neg r(a, b)$, where $r\in \Relation$, $a$ and $b$ can be either a variable $V_\cdot$ or an entity in $\Vertex$.


\subsection{Answer Queries with Set Operator Trees}
Queries that can be answered by set operators are of particular interest~\citep{Ren2020BetaEmbeddings}. The answers can be derived by executing set operators in bottom-up order. The leaves of each operator tree are known entities, which are regarded as sets with a single element. The input and the output of each set operator are all sets. We note that queries solvable by set operators are only a fragment of the first-order queries due to their additional assumptions that guarantee their conversion to operation trees~\citep{Wang2021BenchmarkingCombinatorial}. Moreover, the choice of set operators is not unique to representing the entire class. In this work, we focus on the following operators:
\begin{compactdesc}
\item[Set Projections] Derived from the relations.
\item[Set Operations] :~
\begin{compactdesc}
\item[Set Intersection] Derived from conjunction.
\item[Set Union] Derived from disjunction.
\item[Set Complement] Derived from negation.    
\end{compactdesc}
\end{compactdesc}


\subsection{Wasserstein-Fisher-Rao (WFR) Metric}



Wasserstein-Fisher-Rao metric defines the distances between two measures~\citep{chizat2018interpolating}. Consider two discrete measures in $\real^d$, i.e., $\mu = \sum_{i=1}^M u_i\delta_{x_i}$ and $\nu = \sum_{j=1}^N v_i\delta_{y_j}$, where $\delta$ is the Dirac function, $u_i, v_j \geq 0$, and $x_i, y_j\in \real^d$ are the corresponding coordinates for $1\leq i \leq M$ and $1\leq j \leq N$. For short hand, $\mathbf{u} = [u_1, ..., u_M]^\top$ and $\mathbf{v} = [v_1, ..., v_N]^\top$ denote column mass vectors. Then the WFR metric is defined by solving the following minimization problem.
\begin{align}
    \wfr(\mu, \nu; \eta) = \min_{P\in\real^{M\times N}_+} J(P; \mu, \nu, \eta),\label{eq:wfr-definition}
\end{align}
where $P\in \real^{M\times N}$ is the transport plan and $P_{ij}$ indicates the mass transported from $x_i$ to $y_j$.
We denote the global minima $P^*$ of the Problem~\eqref{eq:wfr-definition} as the WFR optimal transport plan.
The objective function reads,
\begin{align}
J(P; \mu, \nu, \eta) = & \sum_{i=1}^M\sum_{j=1}^N C_{ij} P_{ij}\\ & + \mathcal{D}( P\mathbbm{1}_N \| \mathbf{u}) + \mathcal{D}(P^\top\mathbbm{1}_M \| \mathbf{v}), \nonumber 
\end{align}
where $\mathbbm{1}_N$ is the column vector in $\real^N$ of all one elements, and $\mathcal{D}(\cdot\|\cdot)$ is the KL divergence. $C \in \real_+^{M\times N}$ is the cost matrix and $C_{ij}$ indicates the cost from $x_i$ to $y_j$,
\begin{align}
    C_{ij} = -2 \log\left(\cos_{+} \left( \frac{\pi}{2} \frac{\|x_i-y_j\|}{\eta}\right)\right).
\end{align}
where $\cos_+(x) = \cos(x)$ if $|x|<\pi/2$, otherwise $\cos_+(x) = 0$. $\eta$ is the hyperparameter for the transport radius.

One of the key properties of the WFR metric could be understood by the geodesics in WFR space, as stated in Theorem 4.1 by \citet{chizat2018interpolating}. Specifically, for two mass points at positions $x$ and $y$, the transport only applies when $\|x-y\| < \eta$, such as place $1$ and $2$ in Figure~\ref{fig:local-and-global}, otherwise, only local comparison is counted. We see that the $\eta$ controls the scope of the transport process.

\subsection{Entropic Regularized WFR Solution}
The WFR metric in Equation~\eqref{eq:wfr-definition} can be computed by the Sinkhorn algorithm with an additional entropic regularization term~\citep{chizat2018scaling}. Specifically, one could estimate WFR with the following entropic regularized optimization problem,
\begin{align}
    \min_{P\in \real_+^{M\times N}} J(P; \mu, \nu, \eta) + \overbrace{\epsilon \sum_{ij} P_{ij} \log P_{ij}}^{\text{Entropic Regularization}}. \label{eq:entropic-WFR}
\end{align}
The generalized Sinkhorn algorithm~\citep{chizat2018scaling} solves the unconstraint dual problem of Problem~\eqref{eq:entropic-WFR}, which maximizes the objective
\begin{align}
D_{\epsilon}(\phi,\psi; \mathbf{u}, \mathbf{v}, K_\epsilon) = & \langle 1-\phi, \mathbf{u}\rangle + \langle1-\psi, \mathbf{v}\rangle \label{eq:dual-entropic-wfr} \\ & +\epsilon\langle 1- (\phi \otimes \psi)^{\frac{1}{\epsilon}}, K_\epsilon\rangle, \nonumber
\end{align}
where $K_\epsilon = e^{-\frac{C}{\epsilon}}$ is the kernal matrix, $\phi \in \real^{M}$ and $\psi \in \real^{N}$ are dual variables.
The update procedure of the $(l+1)$-th step of the $j$-th Sinkhorn iteration is
\begin{align}
    \phi^{(l+1)} & \gets \left[ \mathbf{u} \oslash \left(K_\epsilon\psi^{(l)}\right)\right]^{\frac{1}{1+\epsilon}}, \label{eq:sinkhorn-1} \\
    \psi^{(l+1)} & \gets \left[\mathbf{v} \oslash \left(K_\epsilon^\top\phi^{(l+1)}\right)
    \right]^{\frac{1}{1+\epsilon}}. \label{eq:sinkhorn-2}
\end{align}
Let $\phi^*$ and $\psi^*$ be the optimal dual variables obtained from a converged Sinkhorn algorithm. The optimal transport plan is recovered by 
\begin{align}
 P^*={\rm diag}(\phi^*) K_\epsilon {\rm diag}(\psi^*).   \label{eq:from-kernel-to-plan}
\end{align}

We could see that the Sinkhorn algorithm employs the matrix-vector multiplication that costs $O(MN)$ time. In contrast to the Wasserstein metric that can be approximated by 1D sliced-Wasserstein~\citep{carriere2017sliced,kolouri2019generalized} under $O((M+N) \log (M+N))$ time, there is no known sub-quadratic time algorithm for even approximated WFR metric, which hinders its large-scale application. In the next section, we restrict set embeddings to bounded measures in $\real$. We further develop an $O(d)$ algorithm by leveraging the sparse structure of kernel matrix $K_\epsilon$.

\section{Wasserstein-Fisher-Rao Embedding}
The goal of this section is to present how to solve complex queries with set embeddings as the Bounded Measure in $\real$. Let the $S$ be an arbitrary set, including the singleton set $\{e\}$ with a single entity $e$, its embedding is $m[S]$. We denote the collection for all bounded measures as $\bm$. Our discussion begins with the discretization of measure $m[S]\in \bm$ to histogram $m^S\in \bmd$, where $\bmd$ is the collection of bounded histograms with $d$ bars. Then we discuss how to parameterize set operators with embeddings in the $\bmd$ and efficiently compute the scoring function in $\bmd$. Finally, we introduce how to learn set embeddings and operators.

\subsection{Discretize BM1Ds into Histograms}
We discretize each $m[S]\in \bm$ as a histogram on a uniform mesh on $\real$. Without loss of generality, the maximum length of bars in the histogram is one, and the mesh spacing is $\Delta$. 
In this way, each $m[S] = \sum_{i=1}^{d} m_i^{S} \delta_{i\Delta} $, where $m_i^{S} \in [0, 1]$ for $1\leq i \leq d$.
Therefore, it is sufficient to store the discretized mass vector $\mathbf{m}^S = [m_1^{S}, \dots, m_d^{S}] \in \bmd$ because the support set $\{i\Delta\}_{i=1}^d$ is fixed for all $m[S]\in \bm$.
Then we discuss set operators on $\bmd$

\subsection{Set Operators on $\bmd$}

\paragraph{Non-parametric Set Operations}
It should be stressed that the mass vector $\mathbf{m}^S\in \bmd$ can be interpreted \textit{locally}, where each element of $\mathbf{m}^S$ is regarded the continuous truth value in fuzzy logic. Therefore, set operations \textbf{intersection} $\cap$, \textbf{union} $\cup$, and \textbf{complement} on the $\bmd$ are modeled by the element-wise $t$-norm on the mass vector $\mathbf{m}^S$. For the $i$-th element of the mass vector, $1\leq i\leq d$,
\begin{align}
 \text{Intersection} \quad    & m^{S_1 \cap S_2}_i = m^{S_1}_i \top  m^{S_2}_i,\\
 \text{Union}  \quad & m^{S_1 \cup S_2}_i = m^{S_1}_i \bot  m^{S_2}_i,\\
 \text{Complement}  \quad  & m^{S^C}_i = 1 - m^{S}_i,
\end{align}
where $\top$ is a $t$-norm and $\bot$ is the corresponding $t$-conorm.

\paragraph{Neural Set Projections}
Each set \textbf{projection} is modeled as functions from one mass vector to another given a relation $r$. We adopt base decomposition~\citep{Schlichtkrull2018ModelingRelational} to define a Multi-Layer Perceptron (MLP) from $[0, 1]^d$ to $[0, 1]^d$. For each fully-connected layer with input $\mathbf{m}^{S,(l)}\in [0, 1]^{d_l}$, the output $\mathbf{m}^{S,(l+1)}\in [0, 1]^{d_{l+1}}$ through relation $r$ is computed by
\begin{align}
    \mathbf{m}^{S,(l+1)} = \sigma(W_r^{(l)}\mathbf{m}^{S,(l)} + b_r^{(l)}),
\end{align}
where $\sigma$ is an activation function, and $W_r^{(l)}$ and $b_r^{(l)}$ are the weight matrix and bias vector for relation $r$ at the $l$-th layer. Specifically,
\begin{align}
 W_r^{(l)} = \sum_{j=1}^K V_j^{(l)} r_j, \quad b_r^{(l)} = \sum_{j=1}^K a_j^{(l)} r_j.
\end{align}
$K$ is the number of bases, $\mathbf{r}\in \real^K$ is the relation embedding. $V_j^{(l)}\in \real^{d_{l+1}\times d_l}$ and $a_j^{(l)}\in \real^{d_{l+1}}$ the are the base weight matrices and base bias vectors at the $l$-th layer, respectively.

\paragraph{Dropout on Set Complement}
Inspired by the dropout for neural networks that improves the generalizability, we propose to apply dropout to the set complement operation. The idea is to randomly alter the elements in mass vectors before the complement operation by randomly setting their values to $\frac{1}{2}$. In this way, the complemented elements are also $\frac{1}{2}$. This technique improves the generalizability of the set complement operator.

\subsection{Scoring function for $\bmd$}
Consider $\mathbf{m}^{S_1}, \mathbf{m}^{S_2} \in \bmd$. It is straight forward to score this pair by $\wfr(\mathbf{m}^{S_1}, \mathbf{m}^{S_2}; \eta)$. However, direct applying the Sinkhorn algorithm requires a $O(d^2)$ time, which hinders the large-scale computation of the WFR metric. In this part, we introduce (1) convolution-based Sinkhorn to reduce the complexity within $O(d)$ time and 
(2) block diagonal transport as an additional mechanism for the \textit{local} and \textit{global} tradeoff besides the transport radius $\eta$. We note that our contribution does not coincide with the recent linear-time ``fast'' Sinkhorn algorithms~\citep{liao2022fast1,liao2022fast2}, which do not apply to unbalanced optimal transport in $\bmd$.

\paragraph{Convolution-based Sinkhorn}
The computational bottleneck for the Sinkhorn update shown in Equation~\eqref{eq:sinkhorn-1} and~\eqref{eq:sinkhorn-2} is the matrix-vector multiplication. When comparing the discretized measures in $\bmd$, $K_\epsilon$ exhibits a symmetric and diagonal structure.
\begin{align}
    K_{\epsilon, ij} = \left\{\begin{array}{cc}
        \cos\left(\frac{\pi}{2}\frac{|i-j|}{\eta / \Delta}  \right)^{\frac{2}{\epsilon}} & |i-j| < \frac{\eta}{\Delta} \\
        0 & \text{o.w.}
    \end{array}\right.
\end{align}
Let $\omega = \lfloor \frac{\eta}{\Delta} \rfloor$ be the window size, the matrix-vector multiplication $K_\epsilon \mathbf{v} = K_\epsilon^\top \mathbf{v}$ could be simplified as a discrete convolution $H(\beta, \omega) * \mathbf{v}$, where the kernel $[H(\beta, \omega)]_k = \cos\left(\frac{\pi\beta}{2\omega} k \right), -\omega \leq k \leq \omega$ and $\beta := \lfloor \frac{\eta}{\Delta} \rfloor / \frac{\eta}{\Delta} \in (1 - \frac{1}{\omega+1}, 1]$ is another hyperparameter. Specifically, the $i$-th element of $H * \mathbf{v}$ is
\begin{align}
    [H(\beta, \omega) * \mathbf{v}]_{i} = \sum_{k = -\omega}^{+\omega} H_k v_{i+k} \mathbf{1}_{1 \leq i+k \leq d},
\end{align}
where $\mathbf{1}_{1 \leq i+k \leq d} = 1$ if and only if $1 \leq i+k \leq d$.
Then the Sinkhorn algorithm could be simplified as
\begin{align}
    \phi^{(l+1)} & \gets \left[ \mathbf{u} \oslash \left(H(\beta, \omega) *\psi^{(l)}\right)\right]^{\frac{1}{1+\epsilon}}, \label{eq:conv-sinkhorn-1} \\
    \psi^{(l+1)} & \gets \left[\mathbf{v} \oslash \left(H(\beta, \omega) *\phi^{(l+1)}\right)
    \right]^{\frac{1}{1+\epsilon}}. \label{eq:conv-sinkhorn-2}
\end{align}
Hence, the time complexity of the Sinkhorn algorithm could be reduced to $O(\omega d)$. In our setting, $\omega$ is the window size that interpolates the global transport and local comparison, and $\beta$ is chosen to be 1 in every setting.

Once the convolution-based Sinkhorn algorithm converged, we could approximate the WFR metric via the $D_\epsilon$ with optimal $\phi^*$ and $\psi^*$.
For complex query-answering, the final answers are ranked by their distances (the smaller, the better). This process could be accelerated by the primal-dual pruning for WFR-based $k$-nearest neighbors~\citep{wang2020robust} or the Wasserstein Dictionary Learning~\citep{schmitz2018wasserstein}.

\begin{figure}[t]
    \centering
    \includegraphics[width=\linewidth]{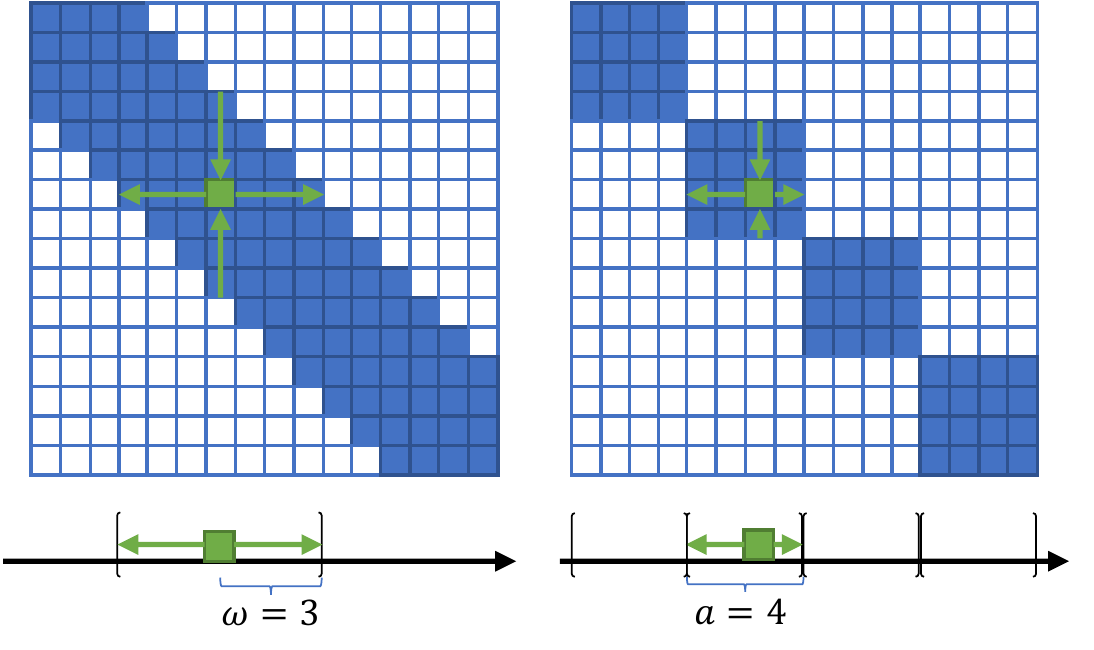}
    \caption{Example of $16\times 16$ transport plan matrices by two mechanisms. The zero elements are indicated by white blocks while the (possible) non-zero elements are colored.  The transport scope of a sample mass point (green block) is illustrated by the arrows. Left: \textit{Relative} scope by the WFR transport of window size $\omega=3$; Right: \textit{Absolute} scope by the block diagonal kernel of block size $a=4$.}
    \label{fig:block-diagonal}
\end{figure}

\paragraph{Block Diagonal Transport}
Besides the window size $\omega$ that controls the scope of transport \textit{relative} to each mass point, we provide another mechanism to restrict the scope of the transport by the \textit{absolute} position of each mass point. Specifically, we consider the block diagonal kernel matrix $K_\epsilon^{b}$ of $b$ blocks, and  $a = d/b$ is the size of each diagonal block. We could see from Equation~\eqref{eq:from-kernel-to-plan} that the block diagonal kernel leads to the block diagonal transport plan. Figure~\ref{fig:block-diagonal} illustrates the differences between the two mechanisms for restricting global transport in terms of possible transport plans.

\paragraph{Computing the Scoring Function}
We propose to define the scoring function ${\rm Dist}$ computed by a convolution-based Sinkhorn with a block diagonal kernel.
It should be stressed that a Problem~\eqref{eq:entropic-WFR} of size $d\times d$ could be regarded as solving $b$ independent problems of size $a\times a$ under the block diagonal problem. This behavior encourages a greater parallelization of the Sinkhorn iterations~\eqref{eq:conv-sinkhorn-1} and~\eqref{eq:conv-sinkhorn-2}. We assume $a > \omega$ to ensure each block contains at least a window size of WFR transport so that those two mechanisms could work together. Given the parallel nature of 1D convolution, the entire distance can be highly parallelized with GPU. Specifically, the scoring function ${\rm Dist}$ is given in Algorithm~\ref{algo:score}.

\begin{algorithm}[t]
\caption{Scoring function on $\bmd$ (PyTorch-like style)}\label{algo:score}
\begin{algorithmic}[1]
\Require two bounded measures $\mathbf{m}^{S_1}, \mathbf{m}^{S_2} \in \bmd$,  entropic regularization $\epsilon$, window size $\omega$, number of blocks $b$ such that the block size $a = d/b \geq \omega$, number of iteration $L$.
\Procedure{\rm Dist}{$\mathbf{m}^{S_1}, \mathbf{m}^{S_2},  \epsilon, \omega, a, b, L$}

\State $ M_1\gets \mathbf{m}^{S_1}\texttt{.reshape}(1,b,a)$.
\State $ M_2\gets \mathbf{m}^{S_2}\texttt{.reshape}(1,b,a)$.
\State Initialize $H \gets H(1, \omega)$.
\State Initialize $\psi \gets \texttt{ones}(1,b,a)$.
\For{$l = 1, ..., L$}
\State $\phi\gets M_1 / \texttt{conv1d}(\psi, H)$.
\State $\psi\gets M_2 / \texttt{conv1d}(\phi, H)$.
\EndFor
\State $\phi^*\gets \phi\texttt{.reshape}(d))$
\State $\psi^*\gets \psi\texttt{.reshape}(d))$
\State \Return $D_\epsilon(\phi^*, \psi^*; \mathbf{m}^{S_1}, \mathbf{m}^{S_2}, K_\epsilon^b)$.
\EndProcedure
\end{algorithmic}
\end{algorithm}

\subsection{Learning Embeddings in $\bmd$}


Let $\mathbf{m}^Q\in \bmd$ be a query embedding of query $Q[V_?]$ and $\mathbf{m}^e\in \bmd$ be the set embedding for unitary set $\{e\}$ with element $e$.
We follow the practice in~\citet{Ren2020BetaEmbeddings} to train the parameterized projections and embeddings with negative sampling. For a query $Q$, we sample one answer $a$ and $K_{\rm neg}$ negative samples $\{v_k\}_{k=1}^{K_{\rm neg}}$. The objective function is
\begin{align}
    L =& -\log \sigma\left(\gamma - \rho{\rm Dist}\left(\mathbf{m}^a,\mathbf{m}^{Q}\right)\right) \\ &-\sum_{k=1}^{K_{\rm neg}} \frac{1}{K_{\rm neg}} \log \sigma\left(\rho{\rm Dist}\left(\mathbf{m}^{v_k},\mathbf{m}^{Q}\right) - \gamma\right), \nonumber
\end{align}
where $\gamma$ is the margin, and $\rho$ is the scale, and $\sigma$ is the sigmoid function.

\begin{table*}
\centering
\tiny
\caption{MRR scores for answering all tasks on FB15k, FB15k-237, and NELL. Scores of baselines are taken from their original paper. The boldface indicates the best scores. $A_{\rm P}$ is the average score for queries without negation (EPFO queries). $A_{\rm N}$ is the average score for queries with negation.}
\label{tab:standard}
\begin{tabular}{llrrrrrrrrrrrrrrrr}
    \toprule
    Dataset & QE & 1P & 2P & 3P & 2I & 3I & PI & IP & 2U & UP & 2IN & 3IN & INP & PIN & PNI & $A_{\rm P}$ & $A_{\rm N}$\\ \midrule
    \multirow{5}{*}{FB15k} &BetaE & 65.1& 25.7  & 24.7 & 55.8 & 66.5 & 43.9 & 28.1 & 40.1 & 25.2 & 14.3 & 14.7  & 11.5 & 6.5 & 12.4 &41.6 & 11.8 \\
    & ConE& 75.3 & 33.8 & 29.2 & 64.4 & 73.7 & 50.9 & 35.7 & 55.7 & 31.4 & 17.9 & 18.7 & 12.5 & 9.8 & 15.1 &49.8 & 14.8 \\
    & MLPMIX & 69.7 & 27.7 & 23.9 & 58.7 & 69.9 & 46.7 & 30.8& 38.2 & 24.8  & 17.2 & 17.8  & 13.5 & 9.1 & 15.2 &  43.4 & 14.8\\
    & Q2P   & 82.6          & 30.8          & 25.5          & 65.1          & 74.7          & 49.5          & 34.9          & 32.1          & 26.2          & 21.9          & 20.8          & 12.5          & 8.9           & 17.1          & 46.8          & 16.4 \\
    & GammaE & 76.5 & 36.9 & \textbf{31.4} & 65.4 & 75.1 & 53.9 & 39.7 & \textbf{53.5} & 30.9 & 20.1 & 20.5 & 13.5 & 11.8 & 17.1 & 51.3 &16.6\\
    & WFRE & \textbf{81.1} & \textbf{37.7} & 30.5 & \textbf{68.5} & \textbf{78.0} & \textbf{56.3} & \textbf{41.8} &48.0 & \textbf{33.1} & \textbf{26.1} & \textbf{26.5}& \textbf{15.6} & \textbf{13.7} & \textbf{19.4} & \textbf{52.8} & \textbf{20.2}\\
    \midrule
    \multirow{6}{*}{FB15k-237} &BetaE & 39.0& 10.9  & 10.0 & 28.8 & 42.5 & 22.4 & 12.6 & 12.4 & 9.7 & 5.1& 7.9  & 7.4 & 3.6 & 3.4 &20.9 &5.4  \\
    & ConE& 41.8 & 12.8 & 11.0 & 32.6 & 47.3 & 25.5 & 14.0 & 14.5 & 10.8 & 5.4 & 8.6 & 7.8 & 4.0 & 3.6 & 23.4 & 5.9\\
    & MLPMIX & 42.4 & 11.5 & 9.9 & 33.5 & 46.8 & 25.4 &14.0 & 14.0 & 9.2  & 6.6 & 10.7  & 8.1 & 4.7 & 4.4 & 22.9  & 6.9 \\
    & Q2P           & 39.1          & 11.4          & 10.1          & 32.3          & 47.7          & 24.0          & 14.3          & 8.7           & 9.1           & 4.4           & 9.7           & 7.5           & 4.6 & 3.8           & 21.9          & 6.0 \\
    & GammaE& 43.2 & 13.2 & 11.0 & 33.5 & 47.9 & 27.2 & 15.9& 13.9 & 10.3 & 6.7 & 9.4 & \textbf{8.6} & 4.8 & \textbf{4.4} & 24.0 & 6.8 \\
    & WFRE & \textbf{44.1} & \textbf{13.4} & \textbf{11.1} & \textbf{35.1} & \textbf{50.1} & \textbf{27.4} & \textbf{17.2} & \textbf{13.9} & \textbf{10.9} & \textbf{6.9} & \textbf{11.2} & \textbf{8.5} & \textbf{5.0} & 4.3 & \textbf{24.8} & \textbf{7.2}\\
    \midrule
    \multirow{6}{*}{NELL} 
    & BetaE & 53.0 & 13.0 & 11.5 & 37.6 & 47.5 & 24.1 & 14.3 & 12.2 & 8.5 & 5.1 & 7.8  & 10.0 & 3.1 & 3.5 & 24.6 & 5.9\\
    & ConE & 53.1 & 16.1 & 13.9 & 40.0 & 50.8 & 26.3& 17.5 & 15.3 & 11.3  & 5.7 & 8.1 & 10.8 & 3.5 & 3.9 & 27.2 & 6.4 \\
    & MLPMIX & 55.4 & 16.5 & 13.9 & 39.5 & 51.0 & 25.7 & 18.3 & 14.7 & 11.2 & 5.1 & 8.0 & 10.0 & 3.1 & 3.5 & 27.4 & 5.9\\  
    & Q2P           & 56.5          & 15.2          & 12.5          & 35.8          & 48.7          & 22.6          & 16.1          & 11.1          & 10.4          & 5.1           & 7.4           & 10.2          & 3.3          & 3.4           & 25.5          & 6.0           \\
    & GammaE & 55.1 & 17.3 & 14.2 & \textbf{41.9} & 51.1 & 26.9& 18.3 & 15.1 & 11.2 & 6.3 & 8.7 & 11.4 & 4.0 & \textbf{4.5} & 27.9 & 7.0\\
    & WFRE & \textbf{58.6}& \textbf{18.6} & \textbf{16.0} & 41.2 & \textbf{52.7} & \textbf{28.4} & \textbf{20.7} & \textbf{16.1}  & \textbf{13.2} & \textbf{6.9} & \textbf{8.8}  & \textbf{12.5}  & \textbf{4.1} & 4.4 & \textbf{29.5} & \textbf{7.3}\\
    \bottomrule
    \end{tabular}
\end{table*}

\section{Experiments}

In this section, we evaluate the performance of WFRE on complex query answering in three aspects: (1) we compare WFRE with other SOTA query embedding methods over commonly used datasets on three knowledge graphs~\citep{Ren2020BetaEmbeddings}; (2) we evaluate WFRE on 301 query types to justify its combinatorial generalizability~\citep{Wang2021BenchmarkingCombinatorial}; (3) we train and evaluate WFRE on a complex query answering datasets on WordNet~\citep{Miller1995WordNetlexical}, a lexical KG whose relations are typically hierarchical~\citep{Huang2022LinELogical}. 
Aspects (2) and (3) emphasize on different query types and the underlying KG, respectively. These results provide empirical evidence for WFRE's strong capability for applying to various query types and KGs.
Moreover, we also investigate the \textit{local} and \textit{global} tradeoff of WFRE on $\omega$ and $a$ in the ablation study. Other results are listed in the Appendix.

\subsection{Experimental Settings}
For all experiments, we follow the practice of training and evaluation in~\citet{Ren2020BetaEmbeddings}. We train query embeddings on train data, select hyperparameters on valid data, and report the scores on test data. Details about the training and evaluation protocol are described in Appendix~\ref{app:train-eval}. For WFRE, the hyperparameters are listed and discussed in Appendix~\ref{app:setting}. All experiments are conducted on one V100 GPU of 32G memory with PyTorch~\citep{NEURIPS2019_9015}.

\subsection{Benchmark Datasets}
Datasets on FB15k-237~\citep{Bordes2013TranslatingEmbeddings}, FB15k~\citep{Toutanova2015Observedlatent}, and NELL~\citep{xiong2017deeppath} proposed by~\citep{Ren2020BetaEmbeddings} are commonly used to evaluate the effectiveness of query embedding methods. WFRE is compared with baselines with local comparison and global transport, including BetaE~\citep{Ren2020BetaEmbeddings}, ConE~\citep{Zhang2021ConECone} MLPMIX~\citep{Alivanistos2022QueryEmbedding}, Q2P~\citep{Bai2022Query2ParticlesKnowledge}, and GammaE~\citep{Yang2022GammaEGamma}. For fairness, we compare the union operators with the DNF treatment introduced by~\citet{Ren2020BetaEmbeddings} where scores of answers are merged from those scores of the containing conjunctive queries. Other treatments about union operators are discussed in Appendix~\ref{app:union} Detailed discussions about the datasets and baselines are listed in Appendix~\ref{app:setting-benchmark}. Table~\ref{tab:standard} shows how WFRE outperforms existing methods by a large margin in terms of the scores averaged from queries with and without logic negation. 




\begin{table*}
\tiny
\centering
    \caption{MRR scores of different query embedding methods on WN18RR. $A_{\rm p}$ is the average of scores from 1P, 2P, and 3P queries; $A_{\ell}$ is the average of scores from other queries without negation; $A_{\rm N}$ is the average of scores from queries with negation. Scores are taken from~\citet{Huang2022LinELogical}.}
    \label{tab:hierarchical-kg}
    \begin{tabular}{lrrrrrrrrrrrrrrrrr}
    \toprule
    QE & 1P & 2P & 3P & 2I & 3I & IP & PI & 2IN & 3IN & INP & PIN & PNI & 2U & UP & $A_{\rm p}$ & $A_{\ell}$ & $A_{\rm N}$
    \\ \midrule 
    BetaE & 44.13 & 9.85 & 3.86 & 57.19 & 76.26 & 17.97 &32.59 &12.77 & 59.98 & 5.07 & 4.04 & 7.48 &7.57 & 5.39 & 19.28 & 32.83 & 17.87 \\ 
    LinE & 45.12 & 12.35 & 6.70 & 47.11 & 67.13 & 14.73 & 24.87 & 12.50 & 60.81 & 7.34 & 5.20 & 7.74 & 8.49 & 6.93 & 21.39 & 28.21 & 18.72 \\ 
    WFRE & 52.78 & 21.00 & 15.18 & 68.23 & 88.15 & 26.50 & 40.97 & 18.99 & 69.07 & 14.29 & 11.06 & 11.01 & 15.14 & 15.81 & 29.65 & 42.46 & 24.88 \\ \bottomrule
    \end{tabular}
\end{table*}

\subsection{Combinatorial Generalization on Queries}

\begin{figure}[t]
    \centering
    \includegraphics[width=\linewidth]{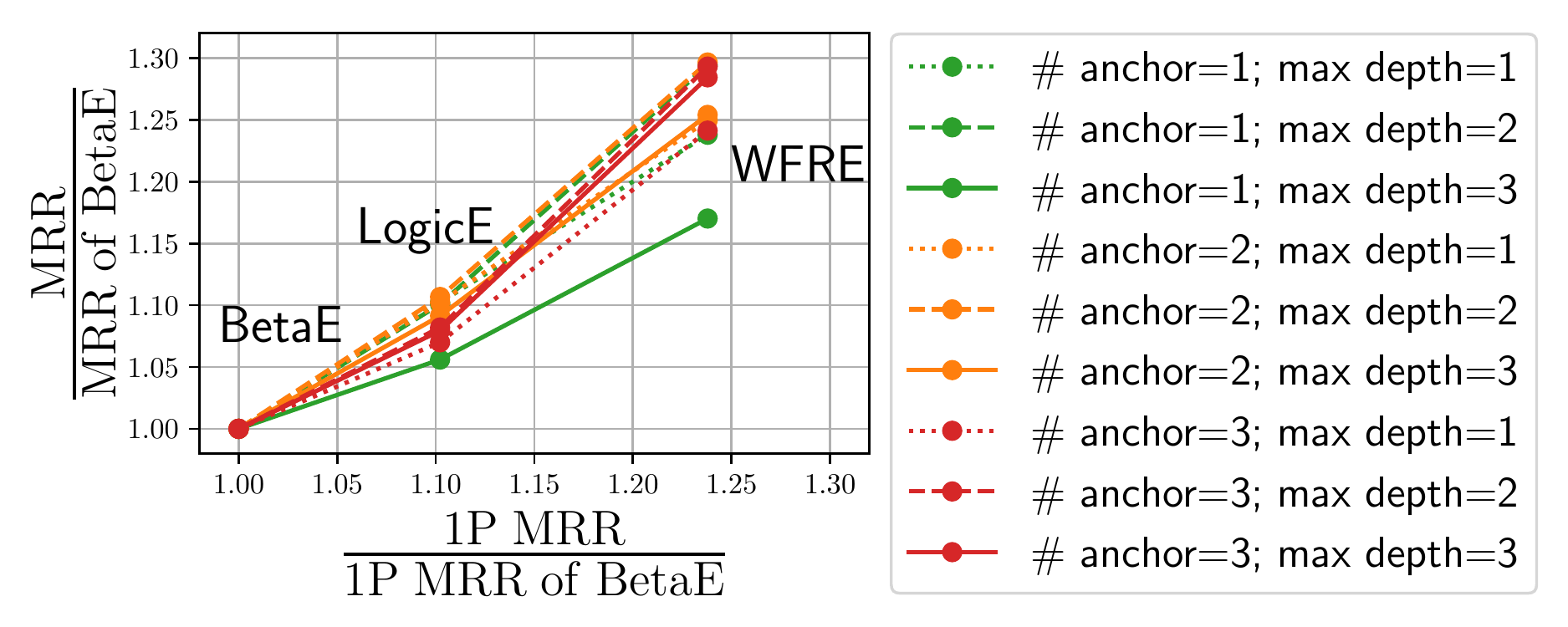}
    \caption{Visualization of different query embedding methods on combinatorial generalizability benchmark~\citep{Wang2021BenchmarkingCombinatorial}. Results of BetaE and LogicE are taken from~\citet{Wang2021BenchmarkingCombinatorial}. The slopes of lines indicate how the performance of a complex query grows as the performance of the one-hop query grows.}
    \label{fig:comb-query}
\end{figure}

We also explore how WFRE generalizes on the combinatorial space of complex queries on a benchmark targeting the combinatorial generalizability of query embedding methods~\citep{Wang2021BenchmarkingCombinatorial}. Details of datasets are presented in Appendix~\ref{app:comb-query}

Results of 301 different query types are averaged by the number of anchor nodes and the maximum depth of the operator tree and are visualized in Figure~\ref{fig:comb-query}.
To illustrate the combinatorial generalizability of complex queries, we normalize scores on query types with the scores on BetaE, as indicated in the axis labels in Figure~\ref{fig:comb-query}. Then we plot the results into lines by the number of anchor nodes and the max depths. Scores from the same model are located at the same vertical line. We find that WFRE not only improves the performance significantly but also generalizes better in combinatorial complex queries with a larger slope compared to LogicE~\citep{Luus2021LogicEmbeddings}.

\subsection{Complex Queries on Hierarchical KG}

Evaluations above are restricted to three commonly used knowledge graphs. Then, we turn to another type of the underlying knowledge graph, which is characterized by the hierarchy of its relation. We train and evaluate WFRE on a complex query dataset proposed by~\citet{Huang2022LinELogical} on WordNet~\citep{Miller1995WordNetlexical}. Details of this dataset are shown in Appendix~\ref{app:hierarchical-kg}. We compare WFRE to LinE~\citep{Huang2022LinELogical}, another histogram-based query embedding proposed to solve queries on hierarchical KG without global transport.
Table~\ref{tab:hierarchical-kg} shows the results on WR18RR. We could see that WFRE significantly outperforms LinE and BetaE.
In particular, WFRE significantly improved the performance of BetaE and LinE on longer multi-hop queries, i.e., 1P,  2P, and 3P queries. It should be stressed that LinE also used histograms as WFRE but trained with the scoring function motivated only by local comparison. This result shows that WFRE is suitable for modeling hierarchical relations because the \textit{local} and \textit{global} tradeoff on the scoring function learns better embeddings WFRE. It also confirms that Wasserstein spaces make the embeddings more efficient~\citep{frogner2018learning}.

\subsection{Local and Global Trade-off}
\begin{figure}[t]
    \centering
    \includegraphics[width=\linewidth]{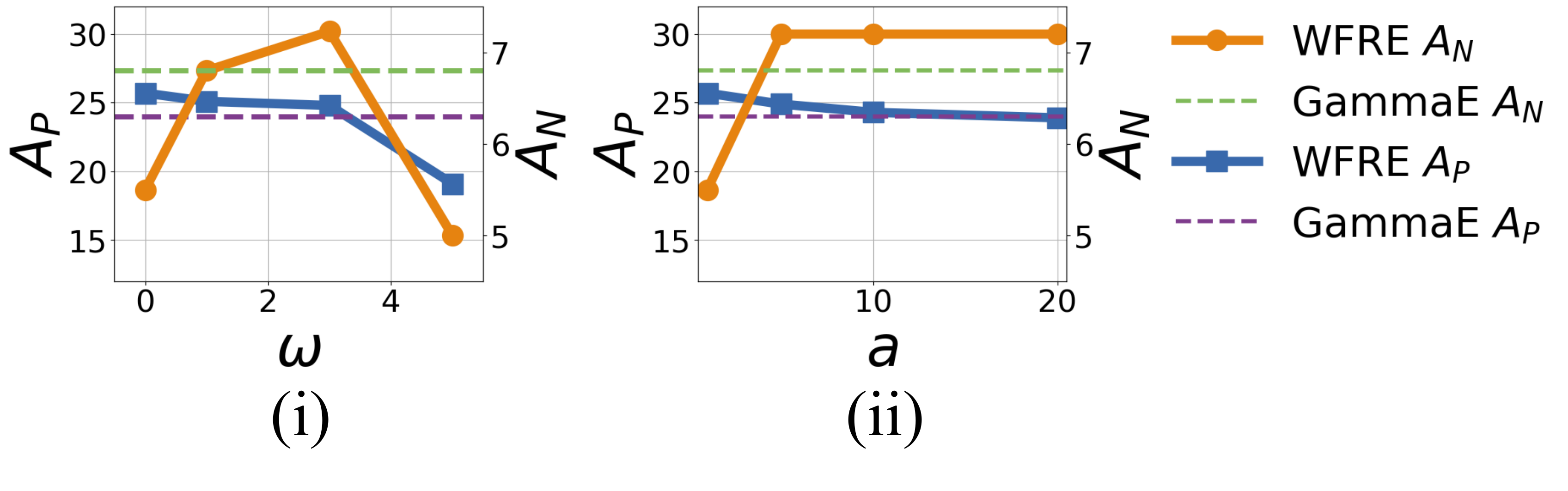}
    \caption{The effect of hyperparameter $\omega$ and $a$. The default choice is $(\omega, a) = (3, 5)$.}
    \label{fig:local-and-global-experiments}
\end{figure}

We further investigate how two mechanisms to restrict the transport, i.e., transport window size $\omega$ and block size $a$ affect the performance.
Experiments are conducted on queries on FB15k-237 sampled by~\citet{Ren2020BetaEmbeddings}.
We alter one value and fix another one. The default choice is $(\omega, a) = (3, 5)$.
Figure~\ref{fig:local-and-global-experiments} demonstrates the effect of these two hyperparameters.

Compared to the most recent SOTA query embedding GammaE~\citep{Yang2022GammaEGamma}, the result confirms the importance of the trade-off between local comparison and global transport. When the block size $a=5$, we find that larger window size $\omega$ hurts the performance of negation.
Meanwhile, the performance of queries without negation (EPFO queries) reaches their maximum when properly choosing $\omega=3$.
When the window size is fixed $\omega = 3$ and $a$ is small, we see that the performance of EPFO and negation queries follows our observation for window size. Further increasing the block size $a$ only has little impact on the EPFO queries but also hurts the performance of negation queries.
It indicates that a proper $a$ is necessary for performance when $\omega$ is fixed. This observation could help to improve the degree of parallelization of the convolution-based Sinkhorn algorithm.

\section{Conclusion}
In this paper, we propose WFRE, a new query embedding method for complex queries on knowledge graphs. The key feature of WFRE against to previous methods is its scoring function that balances local comparison and global transport. Empirical results show that WFRE is the state-of-the-art query embedding method for complex query answering, and has good generalizability to combinatorially complex queries and hierarchical knowledge graphs. The ablation study justifies the importance of the local and global trade-off.

\section{Limitation}
WFRE suffers common drawbacks from the existing query embedding methods. The queries that can be solved by such methods are a limited subclass of first-order queries. It is also not clear how to apply WFRE to unseen entities and relations in an inductive setting.

\section{Ethics Statement}
As a query embedding method, WFRE has stronger generalizability to different query types and knowledge graphs.
Experiments and evaluations in this paper involve no ethical issues and are not even related to any human entities.
WFRE could be potentially used to efficiently infer private information from an industrial-level knowledge graph.
This is a common potential risk for approaches targeting data incompleteness and link prediction.

\section*{Acknowledgement}

The authors of this paper were supported by the NSFC Fund (U20B2053) from the NSFC of China, the RIF (R6020-19 and R6021-20) and the GRF (16211520 and 16205322) from RGC of Hong Kong, the MHKJFS (MHP/001/19) from ITC of Hong Kong and the National Key R\&D Program of China (2019YFE0198200) with special thanks to HKMAAC and CUSBLT. We also thank the support from NVIDIA AI Technology Center (NVAITC) and the UGC Research Matching Grants (RMGS20EG01-D, RMGS20CR11, RMGS20CR12, RMGS20EG19, RMGS20EG21, RMGS23CR05, RMGS23EG08).

\bibliography{ref}
\bibliographystyle{acl_natbib}

\clearpage

\appendix

\begin{table*}[t]
\small
\centering
    \caption{Hyperparameters used for WFRE}
    \label{hyper-descrip}
    \begin{tabular}{lll}
    \toprule
    Hyperparameter & Comments  & Choices
    \\ \midrule 
    Learning rate & Model's convergence  & $\{0.0001, 0.0005, 0.001\}$  \\ 
    Training steps & Model's convergence  & $\{240000, 300000, 360000\}$  \\ 
    Negative sample size~$K_{\rm neg}$  & Model's convergence  & $\{32 \}$  \\ 
    Weight decay & Regulararization for model  & $\{0.001, 0.005, 0.01\}$  \\ 
    $\text{Drop}_p$ & Regulararization for projection operation  & $\{0.05\}$  \\ 
    $\text{Drop}_n$ & Regulararization for negation operation & $\{0.05, 0.15, 0.25\}$ \\
 Entity dimmension~$d$ & Representation of entities& $\{400, 800, 1600\}$\\ 
    Number of relation bases~$K$& Representation of relations& $\{70, 90, 120\}$\\  
    Margin $\gamma$ & Loss function& $\{37.5\}$ \\ 
    Scale $\rho$ & Loss function & $\{90, 120, 150\}$ \\ 
    Size of diagonal block~$a$ & Representation of entities & \{5, 10, 20\} \\
    Window size~$\omega$ & Transport area of WFR distance & $\{1, 3, 5\}$\\ 
     SinkHorn's reg $\epsilon$& Entropy regularization of WFR distance & $\{0.1\}$\\  
     Sinkhorn's maximum iteration $K_S$& Sinkhorn algorithm's convergence & $\{10, 15, 30\}$\\  
 \bottomrule
    \end{tabular}
\end{table*}

\begin{table*}[h]
\scriptsize
\centering
    \caption{Best hyperparameters on every dataset}
    \label{hyper-choice}
    \begin{tabular}{lcccccccccccccccc}
    \toprule
    & learning rate & training steps & $K_{\rm neg}$& weight decay & $\text{Drop}_p$ & $\text{Drop}_n$ & $d$ & $K$ & $\gamma$ &$\rho$ & $a$ & $\omega$& $\epsilon$ & $K_S$
    \\ \midrule 
    FB15k& 0.0005 & 360000 & 32& 0.01 & 0.05 & 0.1 &1600 & 90 & 37.5 & 150 &5 & 3 & 0.1 & 10 \\
    FB15k237 & 0.0005 & 240000 & 32& 0.01 & 0.05 & 0.1 &1600 & 120 & 37.5 & 120  & 5 & 3 & 0.1 & 10 \\
    NELL & 0.0005 & 240000 & 32& 0.01 & 0.05 & 0.1 &1600 & 70 & 37.5 & 180 & 5 & 3 & 0.1 & 10\\
    WN18RR & 0.001 & 120000 & 32& 0.01 & 0.05 & 0.1 & 800 & 70 & 37.5 & 120  & 5 & 3 & 0.1 & 15 \\
 \bottomrule
    \end{tabular}
\end{table*}

\section{Other Methods for Complex Query Answering}\label{app:other-methods}

Despite of computing query embedding with neural set operators, other approaches are also proposed to derive answers. \citet{Daza2020MessagePassing,Liu2022MaskReason} explored the graph representation to answer the logical queries with graph neural networks while \citet{Kotnis2021AnsweringComplex} discussed the logical queries as sequence representation. \citet{Arakelyan2021ComplexQuery} solves the logical queries by solving the continuous optimization problems induced by neural link predictors. However, these discussions are only limited to EPFO queries \textit{without} logical negation. It is not clear how these methods handle first-order queries. 

Meanwhile, neural symbolic methods estimate the probability for whether each entity is the answer set~\citep{Zhu2022NeuralSymbolicModelsa,Xu2022NeuralSymbolicEntangleda} even at each intermediate step. Therefore, it requires $O(|\Vertex| + |\Triple|)$ space and time to derive answers for a given query, where $\Vertex$ and $\Triple$ are the entity set and the triple set of a knowledge graph. Compared to the query embedding methods that require only $O(d)$, where $d$ is the fixed dimension of the embedding space, it is challenging to scale neural symbolic methods to logical queries on large-scale knowledge graphs~\citep{Ren2022SMOREKnowledgea}.

\section{Training and Evaluation Protocal}\label{app:train-eval}
We follow the commonly used experiment settings for EFO-1 query answering, which aims to find non-trivial answers in incomplete graphs and generalize to queries of unseen types.

Given an underlying KG $\mathcal{G} = (\mathcal{V}, \mathcal{R})$ and its triple set $\mathcal{T}$, we sample three subgraphs by change the scope of triples $\mathcal{T}_{\text{train}} \subset \mathcal{T}_{\text{valid}} \subset \mathcal{T}_{\text{test}} = \mathcal{T}$.
Following the standard evaluation protocol,  we aim to find the non-trivial answers which cannot be directly discovered by traversing graphs. 
We denote $[q]_{\text{train}}$ as the answer set of query $q$ in the train graph, the answer set we focus on is $[q]_{\text{test}}\backslash [q]_{\text{train}}$, and these are easy answers that can only be reasoned or predicted. The hard answers are $[q]_{\text{test}}~\backslash~[q]_{\text{valid}}$. Then we would rank the easy(hard) answers against all the non-answer sets $\mathcal{V}/[q]_{\text{valid}}$($\mathcal{V}/[q]_{\text{test}}$). After getting the rank $r$, we calculated mean reciprocal rank (MRR): $\frac{1}{r}$ and Hits at K(Hits@K):$1_{r<K}$ as metric to measure the performance of models.

\section{Settings for WFRE} \label{app:setting}

Our framework is implemented with Pytorch. Our code is based on the pipeline for the EFO-1-QA benchmark~\citep{Wang2021BenchmarkingCombinatorial} and we use AdamW as the optimizer.

There are also some hyperparameters in code. We apply dropout on projection network and denote the drop probability as $\text{Drop}_p$. The Sinkhorn's algorithm's maximum iteration is denoted as $K_S$. And We just set the layer of Projection MLP as 1 because of the results of the experiment relsults. The hyperparameters and their related information in WFRE are listed in Table~\ref{hyper-descrip}. We finetune the hyperparameters for four datasets and the results are presented in Table~\ref{hyper-choice}. Hope the two tables could help you quickly understand our model's hyperparameters.

\begin{table*}[t]
 \centering
 \caption{Number of training, validation, and test queries generated for different query structures.}
 \label{tab: number queries}
 \begin{tabular}{lcccccc}
 \toprule
 \multirow{2}{*}{Dataset} & \multicolumn{2}{c}{Training} & \multicolumn{2}{c}{Validaton} & \multicolumn{2}{c}{Test}\\
 &1P/2P/3P/2I/3I & 2IN/3IN/INP/PIN/PNI &1P &Others &1P & Others \\
 \midrule
 FB15k & 273,710 & 27,371 & 59,097 & 8,000 & 67,016 & 8,000  \\
FB15k-237 & 149,689 &14,968 &20,101 &5,000 &22,812 &5,000  \\
NELL995 & 107,982 & 10,798 & 16,927 & 4,000 & 17,034 & 4,000 \\
WN18RR & 103,509 & 10,350 & 5,202 & 1,000 & 5,356 & 1,000  \\
 \bottomrule
\end{tabular}
\end{table*}

\begin{figure}[t]
    \centering
    \includegraphics[width=\linewidth]{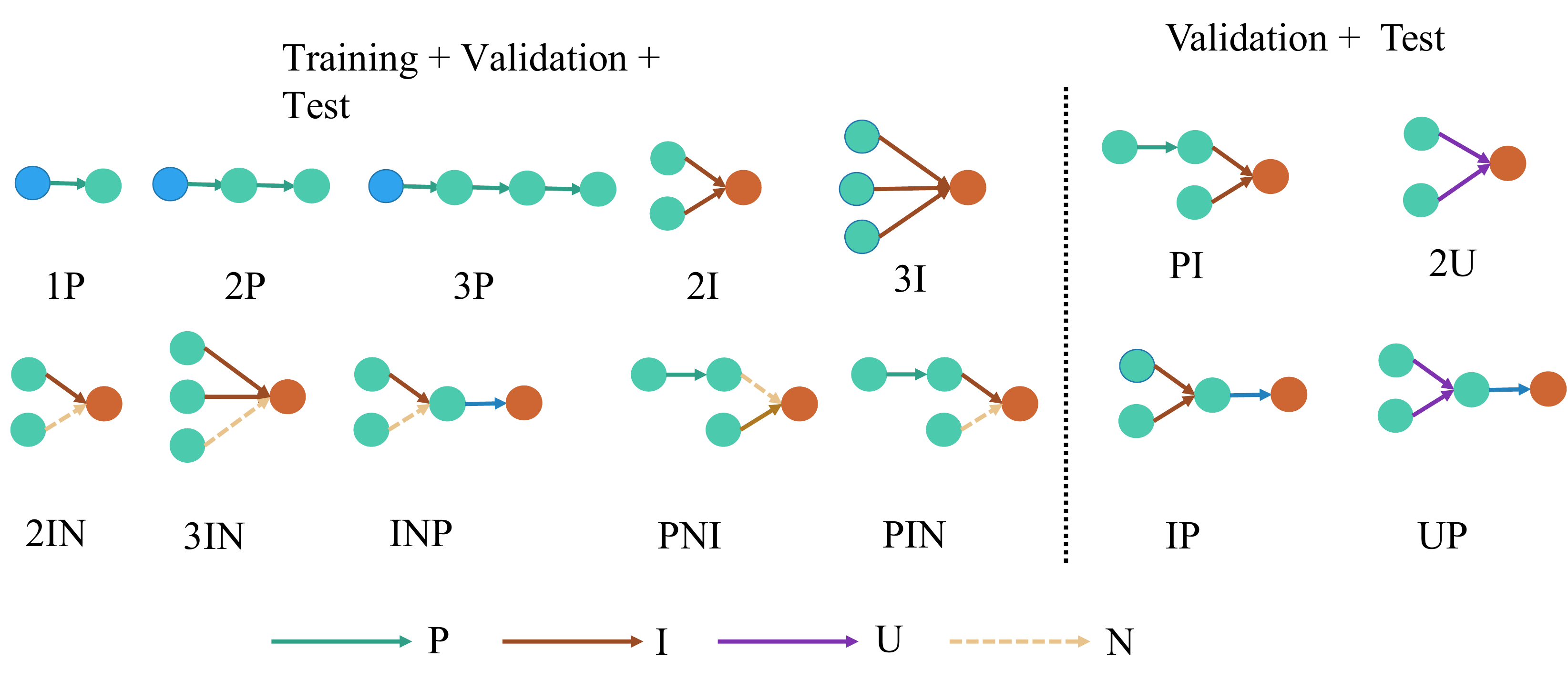}
    \caption{Visualization of logic query structures. The left queries just appear in the training phase, and all the queries are used in the validation and test phases.}
    \label{fig:Visualization-structures}
\end{figure}


\section{Datasets and Baselines}
In this section, we introduce the baselines in three experiments. Table \ref{tab: number queries} presents the basic statistics of different queries on all the benchmark datasets.

\subsection{Benchmark datasets}\label{app:setting-benchmark}
For commonly used dataset~\citep{Ren2020BetaEmbeddings},
there are ten query types 1P, 2P, 3P, 2I, 3I, 2IN, 3IN, INP, PNI, PIN in the training dataset but also four unseen query structures IP, PI, 2U, and UP in the valid and test datasets. The related query structures are visualized in Figure~\ref{fig:Visualization-structures}. The purpose of unseen types of the vaild and test queries is to test the combinatorial generalizability of the neural set operator. 

In this part, we choose the following complex query embedding methods which support arbitrary EFO1 queries:
\begin{compactdesc}
    \item[BetaE~\citep{Ren2020BetaEmbeddings}] Beta distribution embedding whose scoring function is based on local comparison with KL divergence 
    \item[GammaE~\citep{Yang2022GammaEGamma}] GammaE distribution embedding whose scoring function is based on local comparison with KL divergence.
    \item[ConE~\citep{Zhang2021ConECone}] 2D cone embedding whose scoring function is based on global transport with the rotational distance between cones.
    \item[MLP-MIX~\citep{Amayuelas2022NeuralMethods}] Vector embedding whose scoring function is based on local comparison with the Euclidean distance.
    \item[Q2P~\citep{Bai2022Query2ParticlesKnowledge}] Multi-particle embedding whose scoring function is based on global transport by comparing the target particle with the closest particle of a point cloud.
\end{compactdesc}
Those scores are directly taken from corresponding papers. 

\subsection{Combinational generalizability on queries}\label{app:comb-query}
\citet{Wang2021BenchmarkingCombinatorial} propose a new dataset including 301 different query types to benchmark the combinational generalizability of CQA models. Based on the EFO-1 queries represented by OpsTree, EFO-1 formulas are generated with operations including entity, projection, intersection, union, and negation. To make queries more realistic, the maximum length of projection/negation chains and the number of anchor nodes are both limited to no more than 3. The baselines are BetaE and LogicE~\citep{Luus2021LogicEmbeddings}. Scores are directly taken from~\citet{Wang2021BenchmarkingCombinatorial}.

\begin{table*}[t]
    \caption{Hierarchical relations in WN18RR}
    \centering
    \label{tab:hierarchical-relations}
    \begin{tabular}{llllc}
    \toprule
    &Relation & $Khs_{\mathcal{G}_r}$ & $\xi_{\mathcal{G}_r}$ & Hierarchical \\ \midrule 
    & memberMeronym &1.00 & -2.90 & $\checkmark$ \\
    & hypernym & 1.00 & -2.46 & $\checkmark$\\
    & hasPart & 1.00 & -0.82 & $\checkmark$ \\
    & instance hypernym & 1.00 & -0.78 & $\checkmark$ \\
    & memberOfDomainRegion & 1.00 & -0.78 & $\checkmark$ \\
& memberOfDomainUsage & 1.00 & -0.78 & $\checkmark$ \\
& synsetDomainTopicOf & 0.99 & -0.69 & $\checkmark$ \\
& alsoSee & 0.36 & -0.29 & $\times$ \\
& derivationally related form & 0.07 & -3.84 & $\times$ \\
& SimilarTo & 0.07 & -1.00 & $\times$ \\
& verb group & 0.07 & -0.50 & $\times$ \\ \bottomrule
    \end{tabular}
\end{table*}

\subsection{Complex queries on Hierarchical KG}  \label{app:hierarchical-kg}

WN18RR is first introduced as a link prediction dataset created from WN18~\citep{BordesUGWY13}, which is a subset of WordNet. There are 93,003 triples with 40,943 entities and 11 relation types in WN18RR and most of the relations are hierarchical. In Table \ref{tab:hierarchical-relations}, we could know seven out of eleven relations have high antisymmetry $Khs_{\mathcal{G}_r}$~\citet{Huang2022LinELogical} and negative transitive score $\xi_{\mathcal{G}_r}$ \citep{Gu2018LearningMR} and are regarded as hierarchical relations. \citet{Huang2022LinELogical} extends complex logic queries to WN18RR and detaied queries stastics is in Table \ref{tab: number queries}. \citet{Huang2022LinELogical} generated 14 types of queries from hierarchical KG WN18RR and aimed to investigate the reasoning ability of query embeddings in hierarchical knowledge graphs. We choose BetaE and LinE~\citep{Huang2022LinELogical} as baselines, their scores are also taken from~\citet{Huang2022LinELogical}. Notably, LinE~\citep{Huang2022LinELogical} is also a histogram-based query embedding method based on the same closed-form set operation. The key difference between LinE and WFRE is that WFRE encourages the local and global trade-offs.

\section{Modeling Union: DNF and DM}\label{app:union}
There are two ways to deal with union operations. With the De Morgan (DM) Law, it's natural to model union operation $S_1 \cup S_2$ with $ \overline{\overline{S_1} \cap  \overline{S_2}}$. \citep{Ren2020Query2boxReasoning}  transforms queries into a disjunctive normal form (DNF) and only computes the union operation in the last step. Therefore, CQA models usually train intersection and complement logic operations. Though WFRE has closed union operation, WFRE with DNF has better performance as training queries don't contain union operation. 


\section{Addtional results}
Moreover, we further compare with two QE methods FuzzQE~\citep{Chen2022FuzzyLogic} and GammaE~\citep{Yang2022GammaEGamma}. \citet{Yang2022GammaEGamma} develop a new union operation method with the self-attention mechanism and get better performance than DNF and DM. FuzzQE's result on FB15k is missing, and the suggested hyperparameters setting on FB15k-237 is missing. As we couldn't reproduce FuzzQE's result on NELL, we list the results in the paper and those reproduced by us. In Table \ref{tab:add}, WFRE outperforms the two models except for the FuzzQE result in the paper.

Table~\ref{tab:other} also provides the mean and standard derivation of the output of our model. All scores are computed from four runs of cases of different random seeds. We could see that the standard derivation is four orders smaller than the mean value. It shows that WFRE is very stable and significantly outperforms previous baselines.

\begin{table*}
\centering
\tiny
\caption{Additional benchmark comparison on FB15k, FB15k-237, and NELL(MRR).}
\label{tab:add}
\begin{tabular}{llrrrrrrrrrrrrrrrr}
    \toprule
    Dataset & QE & 1P & 2P & 3P & 2I & 3I & PI & IP & 2U & UP & 2IN & 3IN & INP & PIN & PNI & $A_{\rm P}$ & $A_{\rm N}$\\ \midrule
    \multirow{2}{*}{FB15k} 
    & GammaE & 76.5 & 36.9 & 31.4 & 65.4 & 75.1 & 53.9 & 39.7 & 57.1 & 34.5 & 20.1 & 20.5 & 13.5 & 11.8 & 17.1 & 52.3 &16.6\\
    & WFRE & 81.1 & 37.7 & 30.5 & 68.5 & 78.0 & 56.3 & 41.8 &48.0 & 33.1 & 26.1 & 26.5 & 15.6 & 13.7 & 19.4 & 52.8 & 20.2\\
    \midrule
    \multirow{2}{*}{FB15k-237} 
    & GammaE& 43.2 & 13.2 & 11.0 & 33.5 & 47.9 & 27.2 & 15.9& 15.4 & 11.3 & 6.7 & 9.4 & 8.6 & 4.8 & 4.4 & 24.3 & 6.8 \\
    & WFRE & 44.1 & 13.4 & 11.1 & 35.1 & 50.1 & 27.4 & 17.2 & 13.9 & 10.9 & 6.9 & 11.2 & 8.5 & 5.0 & 4.3 & 24.8 & 7.2\\
    \midrule
    \multirow{4}{*}{NELL} 
    & GammaE & 55.1 & 17.3 & 14.2 & 41.9 & 51.1 & 26.9& 18.3 & 16.5 & 12.5 & 6.3 & 8.7 & 11.4 & 4.0 & 4.5 & 28.2 & 7.0\\
    & FuzzQE(our) & 55.5 & 16.8 & 14.4 & 37.3 & 46.9 & 24.0 & 19.1 & 15.0 & 11.7 &7.3 & 9.1 & 11.1 & 4.1 & 4.9 & 26.7 &7.3\\ 
     & FuzzQE(reported) & 58.1 & 19.3 & 15.7 & 39.8 & 50.3 & 28.1 & 21.8 & 17.3    & 13.7 & 8.3 & 10.2 & 11.5 & 4.6 & 5.4 & 29.3 & 8.0\\
    & WFRE & 58.6& 18.6 & 16.0 & 41.2 & 52.7 & 28.4 & 20.7 & 16.1  & 13.2 & 6.9 & 8.8  & 12.5  & 4.1 & 4.4 & 29.5 & 7.3\\
    \bottomrule
    \end{tabular}
\end{table*}

\begin{table*}
\centering
\tiny
\caption{WFRE: metrics' mean values ($\times10^{-2}$) and standard deviations ($\times10^{-6}$, boldface).}
\label{tab:other}
\begin{tabular}{llrrrrrrrrrrrrrrrr}
    \toprule
    Dataset & QE & 1P & 2P & 3P & 2I & 3I & PI & IP & 2U & UP & 2IN & 3IN & INP & PIN & PNI & $A_{\rm P}$ & $A_{\rm N}$\\ \midrule
    \multirow{8}{*}{FB15k} 
    &\multirow{2}{*}{MRR}& 81.1& 37.7 & 30.5 & 68.5 & 78.0 & 56.3 & 41.8 & 48.0 & 33.1 & 26.1 & 26.5 & 15.6 & 13.7 & 19.4 & 52.8 & 20.2 \\
& &  \textbf{0.073} & \textbf{0.33} & \textbf{0.39} & \textbf{0.45} & \textbf{1.5} & \textbf{0.37} & \textbf{1.6} & \textbf{1.3} & \textbf{2.5} & \textbf{0.74} & \textbf{1.2} & \textbf{1.8} & \textbf{0.37} & \textbf{2.8} & \textbf{0.16} & \textbf{0.046}\\\cmidrule{2-18}    
    & \multirow{2}{*}{HITS1}& 73.0 & 27.4 & 21.2 & 58.6 & 70.1 & 45.7 & 31.1 & 36.1 & 23.1 & 16.5 & 16.4 & 8.4 & 7.1 & 10.9 &42.9 & 11.9 \\
& &  \textbf{0.096} & \textbf{0.35} & \textbf{2.0} & \textbf{1.6} & \textbf{5.4} & \textbf{1.0} & \textbf{1.8} & \textbf{2.5} & \textbf{8.1} & \textbf{1.0} & \textbf{1.4} & \textbf{1.3} & \textbf{0.32} & \textbf{4.8} & \textbf{0.67} & \textbf{0.11}\\\cmidrule{2-18}
    & \multirow{2}{*}{HITS3} & 87.8 & 42.0 & 33.8 & 74.5 & 83.5 & 62.0 & 46.6& 54.5 & 36.6  & 28.5 & 28.9  & 16.4 & 13.8 & 20.8 &  58.0 & 21.7\\
& &  \textbf{0.18} & \textbf{1.3} & \textbf{0.35} & \textbf{2.0} & \textbf{0.69} & \textbf{1.9} & \textbf{9.8} & \textbf{3.5} & \textbf{2.6} & \textbf{0.18} & \textbf{6.8} & \textbf{3.0} & \textbf{0.16} & \textbf{1.7} & \textbf{0.11} & \textbf{0.32}\\\cmidrule{2-18}
    & \multirow{2}{*}{HITS10} & 93.5 & 58.0 & 48.5 & 86.5 & 92.2 & 76.3 & 62.4  & 70.4  & 52.7 & 45.5 & 47.3 & 29.9 & 26.6  & 36.2 & 71.2 & 37.1 \\
& &  \textbf{0.38} & \textbf{6.5} & \textbf{1.2} & \textbf{0.60} & \textbf{0.95} & \textbf{0.062} & \textbf{1.1} & \textbf{0.60} & \textbf{2.9} & \textbf{0.97} & \textbf{0.20} & \textbf{0.91} & \textbf{4.3} & \textbf{0.16} & \textbf{0.0086} & \textbf{0.29}\\
    \midrule
    \multirow{8}{*}{FB15k237}
    & \multirow{2}{*}{MRR} & 44.1 & 13.4 & 11.1 & 35.1 & 50.1 & 27.4 & 17.2 & 13.9 & 10.9 & 6.9 & 11.2 &8.5 & 5.0 & 4.3 & 24.8 & 7.2 \\
& &  \textbf{0.032} & \textbf{2.1} & \textbf{2.4} & \textbf{1.0} & \textbf{3.1} & \textbf{1.6} & \textbf{0.36} & \textbf{0.89} & \textbf{0.12} & \textbf{0.60} & \textbf{1.1} & \textbf{0.20} & \textbf{0.42} & \textbf{0.56} & \textbf{0.016} & \textbf{0.17}\\\cmidrule{2-18}
    &\multirow{2}{*}{HITS1} & 33.7 & 7.3 & 5.5 & 23.9 & 39.6 & 18.5 & 10.8 & 7.5 & 5.2 & 2.8 & 5.2 & 3.7 & 1.6 & 1.4 & 16.9 & 2.9\\
& &  \textbf{0.056} & \textbf{5.7} & \textbf{5.7} & \textbf{1.6} & \textbf{5.5} & \textbf{2.1} & \textbf{0.70} & \textbf{2.3} & \textbf{0.37} & \textbf{0.071} & \textbf{1.2} & \textbf{0.76} & \textbf{2.0} & \textbf{0.56} & \textbf{0.15} & \textbf{0.044}\\\cmidrule{2-18}
    & \multirow{2}{*}{HITS3}& 48.9 & 13.8 & 11.3 & 39.7 & 55.2 & 29.9 & 18.0 & 14.0 & 11.0 & 6.3 & 10.8 & 8.2 & 4.4 & 3.6& 27.0 & 6.7\\
& &  \textbf{1.2} & \textbf{0.36} & \textbf{0.92} & \textbf{4.1} & \textbf{4.8} & \textbf{0.69} & \textbf{1.8}  & \textbf{0.11} & \textbf{0.49} & \textbf{3.7} & \textbf{0.42} & \textbf{5.1} & \textbf{0.20} & \textbf{1.1} & \textbf{0.15} & \textbf{0.73}\\\cmidrule{2-18}
    & \multirow{2}{*}{HITS10} & 64.5 & 25.6 & 21.9 & 57.8 & 71.0 & 45.6 &29.8 & 23.1 & 17.5  & 14.4 & 23.1  & 17.5 & 10.9 & 9.0 & 40.5  & 15.0 \\
& &  \textbf{0.15} & \textbf{3.9} & \textbf{1.0} & \textbf{13} & \textbf{1.6} & \textbf{18} & \textbf{0.53}  & \textbf{0.69} & \textbf{0.48} & \textbf{2.5} & \textbf{4.6} & \textbf{0.95} & \textbf{0.37} & \textbf{0.35} & \textbf{0.17} & \textbf{0.24}\\
    \midrule
    \multirow{8}{*}{NELL}
    & \multirow{2}{*}{MRR} & 58.6 & 18.8 & 16.0 & 41.2 & 52.7 & 28.4 & 20.7 & 16.1 & 13.2  & 6.9 & 8.8 & 12.5 & 4.1 & 4.4 & 29.5 & 7.3 \\
& &  \textbf{0.74} & \textbf{0.44} & \textbf{0.056} & \textbf{0.72} & \textbf{6.0}  & \textbf{2.6} & \textbf{0.96}  & \textbf{0.22} & \textbf{0.62} & \textbf{0.0026}& \textbf{0.24} & \textbf{1.6} & \textbf{0.046} & \textbf{0.047} & \textbf{0.52} & \textbf{0.095}\\\cmidrule{2-18}
    & \multirow{2}{*}{HITS1} & 49.1 & 12.6 & 10.6 & 29.5 & 41.4 & 20.6 & 14.1 & 9.4 & 7.8 & 2.3 &3.2  & 6.1 & 1.1 & 1.4  & 21.7 & 2.8\\
& &  \textbf{1.3} & \textbf{0.27} & \textbf{0.40} & \textbf{12} & \textbf{0.77}  & \textbf{5.3} & \textbf{0.34} & \textbf{1.1} & \textbf{0.29}  & \textbf{0.12} & \textbf{0.16} & \textbf{1.9} & \textbf{0.24} & \textbf{0.19} & \textbf{0.44} & \textbf{0.25}\\\cmidrule{2-18}
    & \multirow{2}{*}{HITS3}& 64.2 & 19.9 & 16.8 & 46.5 & 58.3 & 30.8& 22.2 & 17.2 & 13.8 & 6.0 & 7.6 & 12.9 & 3.2 & 3.7 & 32.2 & 6.7\\
& &  \textbf{1.3} & \textbf{8.9} & \textbf{1.6} & \textbf{2.4} & \textbf{2.9} & \textbf{3.5} & \textbf{4.1} & \textbf{5.9} & \textbf{2.1} & \textbf{0.25} & \textbf{0.70} & \textbf{3.2} & \textbf{0.10} & \textbf{0.73} & \textbf{1.71} & \textbf{0.18}\\\cmidrule{2-18}
    & \multirow{2}{*}{HITS10} & 76.1 & 31.0 & 26.3 & 64.6 & 75.0 & 43.9 & 33.7 & 29.4 & 24.0 & 15.6 & 19.7 & 24.8 & 8.6 & 9.1 & 44.5 & 15.5\\
& &  \textbf{1.1} & \textbf{6.2} & \textbf{6.5} & \textbf{5.1} & \textbf{1.6} & \textbf{1.3} & \textbf{6.3} & \textbf{3.0} & \textbf{0.27} & \textbf{0.49} & \textbf{0.062} & \textbf{1.9} & \textbf{1.2} & \textbf{1.5} & \textbf{0.33} & \textbf{0.24}\\
    \bottomrule
    \end{tabular}
\end{table*}

\end{document}